\documentclass[10pt,twocolumn,letterpaper]{article}

\usepackage[pagenumbers]{wacv} 

\usepackage{graphicx}
\usepackage{amsmath}
\usepackage{amssymb}
\usepackage{booktabs}

\usepackage[pagebackref,breaklinks,colorlinks]{hyperref}

\usepackage{multirow}
\usepackage[accsupp]{axessibility}  

\usepackage[capitalize]{cleveref}
\crefname{section}{Sec.}{Secs.}
\Crefname{section}{Section}{Sections}
\Crefname{table}{Table}{Tables}
\crefname{table}{Tab.}{Tabs.}

\newcommand{\stdsize}[1]{\footnotesize #1}
\newcommand{\bdsize}[1]{\normalsize #1}

\let\titleold\title
\renewcommand{\title}[1]{\titleold{#1}\newcommand{\thetitle}{#1}}
\def\maketitlesupplementary
   {
   \newpage
       \twocolumn[
        \centering
        \Large
        \textbf{\thetitle \\ (Supplementary Material)}\\
        \vskip .5em
        \vspace*{12pt}
       ] 
   }

\def\wacvPaperID{} 
\def\confName{WACV}
\def\confYear{2024}

\begin{document}

\title{Self-Supervised Relation Alignment for Scene Graph Generation}
\author{Bicheng Xu$^{1,2}$ \qquad Renjie Liao$^{1,2,3}$ \qquad Leonid Sigal$^{1,2,3}$\\
$^1$University of British Columbia \qquad
$^2$Vector Institute for AI \qquad
$^3$Canada CIFAR AI Chair \\
{\tt\small bichengx@cs.ubc.ca \qquad rjliao@ece.ubc.ca \qquad lsigal@cs.ubc.ca}
}

\maketitle

\begin{abstract}
   The goal of scene graph generation is to predict a graph from an input image, where nodes correspond to identified and localized objects and edges to their corresponding interaction predicates. Existing methods are trained in a fully supervised manner and focus on message passing mechanisms, loss functions, and/or bias mitigation. In this work we introduce a simple-yet-effective self-supervised relational alignment regularization designed to improve the scene graph generation performance. The proposed alignment is general and can be combined with any existing scene graph generation framework, where it is trained alongside the original model's objective. The alignment is achieved through distillation, where an auxiliary relation prediction branch, that mirrors and shares parameters with the supervised counterpart, is designed. In the auxiliary branch, relational input features are partially masked prior to message passing and predicate prediction. The predictions for masked relations are then aligned with the supervised counterparts after the message passing. We illustrate the effectiveness of this self-supervised relational alignment in conjunction with two scene graph generation architectures, SGTR~\cite{li2022sgtr} and Neural Motifs~\cite{zellers2018neural}, and show that in both cases we achieve significantly improved performance. 
\end{abstract}

\section{Introduction}
\label{sec:intro}

Scene graph generation has emerged as a core problem in computer vision in recent years. The task involves producing a graph-based representation of the scene from an image. As the name suggests, scene graph representation, encodes a scene as a graph where {\em nodes} correspond to objects, with corresponding (bounding box) locations and class labels, and directed {\em edges} encode pairwise relations among these objects. The ultimate goal of scene graph generation is to produce such representations from raw images (or videos~\cite{ji2020actiongenome}). Such representations have proved valuable for a variety of higher-level AI tasks, \eg, image/video captioning~\cite{li2019caption,nguyen2021caption}, and visual question answering~\cite{qian2022vqa}. 

\begin{figure}[t]
  \centering
   \includegraphics[width=\linewidth]{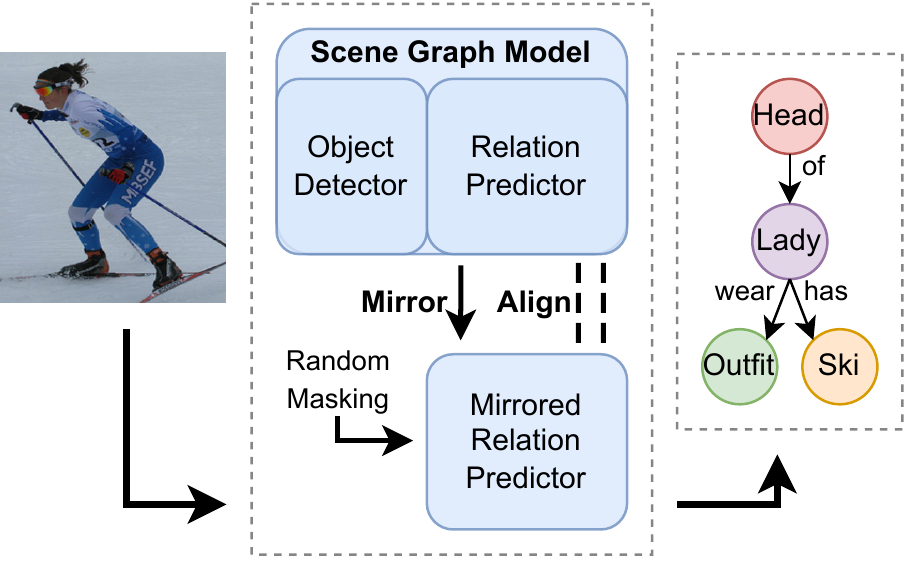}
   \caption{{\bf Illustration of our proposed self-supervised relation alignment mechanism.} A scene graph generation model typically contains an object detector to detect objects of interest, and a relation predictor to predict relations among the detected objects. 
   We create a mirror copy of the original relation predictor, apply random masking on the input to the mirrored relation predictor, and align the relation predictions between the mirrored copy and the original one.}
   \label{fig:teaser}
\end{figure}

Scene graph generation models usually contain two message passing networks: one for detecting objects of interest, and the other for relation prediction among the detected objects. The message passing network can be a recurrent neural network~\cite{tang2019learning,zellers2018neural}, a graph neural network~\cite{yang2018graph,suhail2021energy}, or more recently, a stack of Transformer blocks~\cite{dong2021visual,li2022sgtr,kundu2023ggt}. Newly developed architectures continue to push state-of-the-art in terms of performance. However, one of the fundamental challenges in scene graph generation is sparsity of data, owning to large number of objects, relations and their exponential cross product. Recent approaches have tried to address this by mediating dataset bias in relational classes with either re-sampling~\cite{desai2021learning,li2021bipartite,li2022sgtr} or loss re-weighting~\cite{khandelwal2022iterative} techniques. While this mediates bias among relation predicates, it does not address the core problem of data sparsity, where even for the frequent relations, it is often difficult to observe all appearance variations in objects and their interactions from a dataset like Visual Genome~\cite{krishna2017visual} or similar. 

Equipped with this intuition, we propose a novel self-supervised relational alignment mechanism to alleviate this limitation. Our mechanism is simple, effective and generic, \ie, can be combined with any underlying scene graph generation architecture. Consequently, we illustrate it in conjunction with both a traditional two-stage scene graph generation 
approach -- Neural Motifs~\cite{zellers2018neural} and a recent Transformer-based SGTR~\cite{li2022sgtr}. 
Specifically, given a scene graph generation model with its original objective, we create an auxiliary (mirrored) relation prediction branch which is identical in structure and shares parameters with the main (original) branch, but whose only goal is to align relational predictions with those coming from the main supervised branch, as shown in Figure~\ref{fig:teaser}. 
The {\em alignment} is implemented using KL divergence over predicted predicate label distributions. Importantly, 
In the 
mirrored branch, we apply a random masking on the relational feature input\footnote{The specifics of this depend on the structure of the original scene graph generation model.} prior to the relational message passing / prediction head.
The goal of this masking is to form impoverished augmented views of the data, with the alignment objective driving {\em denoising}, which enhances feature formation and refinement within and across relations. Further, unlike other self-supervised methods that tend to pre-train models using self-supervision and then fine-tune them, we train both the supervised and auxiliary branches jointly.

\vspace{0.1in}
\noindent
{\bf Contributions.} Our contributions are as follows. 
First, we introduce a simple, but effective and general self-supervised alignment mechanism for relation prediction. This alignment is achieved by instantiating an auxiliary branch of the relation prediction network that shares structures and parameters with the original scene graph model (whatever it may be).
In this auxiliary branch we create alternate views of relation features, through random masking, ahead of the message passing. 
The 
corresponding relational predictions are then aligned with the predicate distributions obtained from the  
original relational branch (main branch). 
Second, unlike other self-supervised learning approaches that require two stage training, first self-supervised pre-training and then fine-tuning to the task. We show that we can leverage self-supervised alignment as an effective 
{\em regularization} during training of the main branch, \ie, allowing us to train the two branches jointly and at the same time. 
Third, we illustrate the effectiveness of our self-supervision, implemented via alignment, by pairing it with two very diverse scene graph generation architectures (Neural Motifs~\cite{zellers2018neural} and SGTR~\cite{li2022sgtr}) and showing significant improvements in both cases, without the need to change the original architectures, losses or training in any way. 

\section{Related Work}
\label{sec:related}
\subsection{Scene Graph Generation (SGG)}
Works on scene graph generation can be categorized into two classes: two-stage and one-stage.

{\em Two-stage models} first train a Faster-RCNN~\cite{ren2015faster} (or similar) object detector, and use this trained object detector to obtain image features and bounding box proposals (including their locations 
and feature representations). 
With these as fixed inputs, two-stage approaches then design various message passing networks to refine object and relation features and produce the final scene graph by classifying them. Message passing networks within these approaches can take many forms: LSTMs~\cite{zellers2018neural,tang2019learning}, graph neural networks~\cite{yang2018graph,li2021bipartite}, or others~\cite{lin2020gps,qi2019attentive,xu2017scene,lu2021context,zhang2019graphical,jung2023devil}. From the objective perspective, while most leverage cross-entropy, energy-based losses~\cite{suhail2021energy}, that better capture structure in the output space, have also been proposed.  

{\em One-stage models} are more recent and attempt to directly generate the scene graph from a given image without first pre-training an object region proposal 
network.
These approaches overcome the limitation of the aforementioned two-stage models, where the object proposals are fixed and can not be modified by the scene graph generation model. One-stage models, usually build on one-stage object detectors (\eg, Yolo~\cite{redmon2015yolo} or DETR~\cite{carion2020end}) and employ network structures like fully convolutional networks~\cite{liu2021fully,teng2022structured} or Transformers~\cite{dong2021visual,li2022sgtr,khandelwal2022iterative,cong2023reltr,desai2022single} for predictions.

Our proposed self-supervised relation alignment mechanism is generic, and can be applied to both one-stage and two-stage scene graph generation models. In this paper, we use an one-stage SGTR model~\cite{li2022sgtr} and a two-stage Neural Motifs model~\cite{zellers2018neural}, as examples, to show its effectiveness.

\subsection{Self-Supervision in SGG}
Self-supervised learning is a powerful paradigm to obtain useful feature representations, examples including 
Bert-type models~\cite{devlin2018bert,liu2019roberta, lu2019vilbert,li2019visualbert,su2019vl,bao2021beit}, denoising auto-encoding based approaches~\cite{vincent2010stacked,pathak2016context,he2022masked,xie2022simmim}, and contrastive learning~\cite{chen2020simple,oord2018representation,he2020momentum,tian2020contrastive}. 
These methods typically first define some pretext tasks to train a generic feature extractor, and then utilize the extracted features or fine-tune the pre-trained model to a range of downstream tasks. 

In the scene graph generation literature, self-supervised learning has not been explored much. 
Zareian \etal~\cite{zareian2020learning} use a denoising auto-encoder (DAE) type structure to learn the commonsense that exists in the scene graph labels. Given a scene graph annotation, they mask out some of the object and/or relation annotations, and train a DAE structure to reconstruct these masked out labels. They show that their approach can be applied to any trained scene graph generation model as a post-processing step to correct some predictions which do not follow common sense. 
Hasegawa \etal~\cite{hasegawa2023improving} proposes a scene graph generation model consisting of a pre-trained object detector, a relational encoder, and object and relation classifiers. They first use self-supervision with a contrastive loss to train the relation encoder, and with the relation encoder fixed, they then train the classifiers to generate the scene graph. Their self-supervised technique is specific to their model, and a pre-trained object detector is required.

In contrast to the multi-stage training or post-processing, 
we employ random masking as a type of data augmentation and our self-supervision is achieved via aligning the predictions from the masked feature input with those from the unmasked one. Our introduced self-supervised loss is added to the original loss of the scene graph generation model during training. 

\section{Self-Supervised Relation Alignment}
\label{sec:maskaligh}
In this section, we first describe the general idea behind our proposed self-supervised relation alignment mechanism, and then instantiate it within two popular scene graph generation models: one-staged SGTR~\cite{li2022sgtr} and two-staged Neural Motifs~\cite{zellers2018neural}.

\subsection{General Pipeline}
\begin{figure*}[t]
  \centering
  \includegraphics[width=\linewidth]{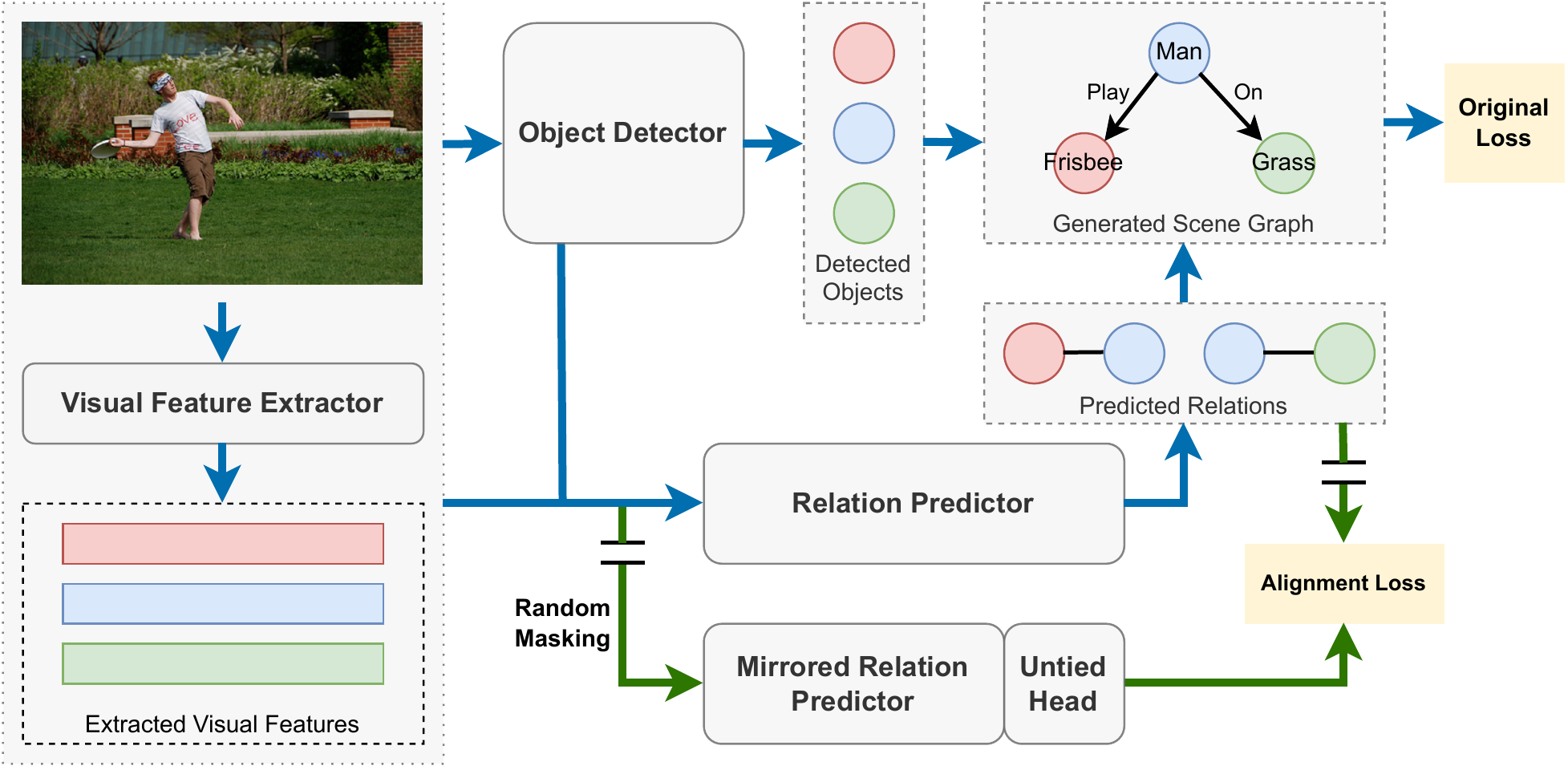}
  \caption{{\bf Our proposed self-supervised relation alignment mechanism.} The blue arrows indicate a standard scene graph generation pipeline, and the green arrows illustrate our proposed relation alignment. The architecture and parameters of Relation and Mirrored Relation Predictors are all {\em shared} except the last relation projection head, which is denoted as {\tt Untied Head}. This untied projection head helps alleviate the burden of performing two tasks (\ie, supervised relation prediction and self-supervised relation alignment) for the original projection head, resulting in better learned representations. In the figure, $=$ indicates the stopping gradient operation.}
  \label{fig:model}
\end{figure*}

Scene graph generation models typically contain three parts: a pre-trained visual feature extractor, an object detector, and a relation predictor. 
As illustrated in Figure~\ref{fig:model}, the feature extractor takes an image as input and computes visual features.
Conditioned on the extracted features, the object detector localizes and classifies the objects of interest in the image.
Finally, the relation predictor predicts the relationship between any pair of detected objects. Depending on the architecture, the relation predictor relies either on the concatenation of features from the pair of objects in question, its own extracted features, or both. 

Motivated by the sparsity of visual data for relations, which stems from the possible exponential space of appearances for interacting objects,  
we focus on building a self-supervised alignment mechanism for the relation predictor. The goal of this alignment is two-fold: (1) to regularize learning with augmented data samples; and (2) to encourage the relation predictor, which includes message passing and refinement of relational features, to more effectively propagate information both within a single relation representation and across such representations.

To build the self-supervised alignment, we first create a duplicate 
of the relation predictor, called \emph{mirrored relation predictor}. 
Note that both the neural network architecture and the weights are shared between the mirrored and the original relation predictors (\ie, the weights are initially the same and updated in the same way at each iteration). 
The input to the mirrored relation predictor is the same as that to the original one, which usually contains image features from the feature extractor and/or object features from the object detector. 
We then apply a random masking (details in Section~\ref{sec:randmask}) on the input to the mirrored relation predictor. 
An alignment loss is introduced to align the relation predictions from the mirrored predictor with those from the original branch. 
To make sure the introduced alignment loss only affects the relation predictor, we stop the gradient of the alignment loss at the input feature. 
This ensures that the alignment loss only affects the message passing and feature refinement in the relation predictor.
The resulting alignment loss is added to the original training loss. 

It is important to mention that the auxiliary branch, in the form of the mirrored relation predictor, is only used during training of the network. 
Once trained, during evaluation, we still only use the original relation predictor to generate the scene graph. In other words, one can view our self-supervised alignment mechanism as a form of (more sophisticated) regularization. The whole pipeline is shown in Figure~\ref{fig:model}. In the following, we describe in detail about how the masking is applied and how the alignment is achieved.

\subsubsection{Random Masking}
\label{sec:randmask}
The original and the mirrored relation predictors both take a feature matrix $\mathbf{R} \in \mathbb{R}^{N \times d}$ as input, where $N$ is the number of relations and $d$ is the dimensionality of feature vector for each relation. The relation predictor typically refines $\mathbf{R}$ (\eg, through a series of LSTM, GNN or Trasformer layers) and uses the refined representations to classify each feature to produce a relation or {\tt no} relation label. 

We apply random masking on the input to the mirrored relation predictor. 
Specifically, for each element $\mathbf{R}_i$ in $\mathbf{R}$, the masking operation independently zeros it out with probability $p$. That is, 
\begin{align}\label{eq:mask_pattern}
\operatorname{masking}(\mathbf{R}_i) &= \left\{ 
                        \begin{array}{ll}
                          0 \;\; &\text{w. probability $p$}, \\
                          \mathbf{R}_i \;\; &\text{w. probability $(1-p)$}. \\
                        \end{array} \right.
\end{align}
We then feed the masked input $\mathbf{R}_{\text{masked}}$ to the mirrored relation predictor, while the original predictor still receives $\mathbf{R}$.

\subsubsection{Self-Supervised Alignment Loss}
\label{subsec:alignloss}
With the masked input $\mathbf{R}_{\text{masked}}$, the mirrored relation predictor generates relation predictions, \ie, the probability distribution over possible relationships, denoted as $P_{\text{masked}}$. 
We align $P_{\text{masked}}$ with the relation predictions from the original relation predictor, denoted as $P_{\text{original}}$. 
We use the Kullback-Leibler (KL) divergence to measure the closeness between $P_{\text{original}}$ and $P_{\text{masked}}$. 
In particular, we fix $P_{\text{original}}$ and treat it as the target in the $\mathrm{KL}$ loss to optimize $P_{\text{masked}}$. 
We term this alignment loss,
\begin{equation}
\mathcal{L}_{\text{align}} = \operatorname{KL}(P_{\text{original}} \Vert P_{\text{masked}}). 
\end{equation}
We add the alignment loss $\mathcal{L}_{\text{align}}$ to the original training loss (typically a weighted sum of cross-entropy losses for object and relation predictions with one-hot class targets)
in the scene graph generation model, denoted as $\mathcal{L}_{\text{original}}$. 
The final training loss can be formulated as
\begin{equation} \label{eq:loss}
\mathcal{L}_{\text{final}} = \mathcal{L}_{\text{original}} + \lambda  \mathcal{L}_{\text{align}},
\end{equation}
where $\lambda$ is a hyper-parameter to control the relative weights between $\mathcal{L}_{\text{original}}$ and $\mathcal{L}_{\text{align}}$ .

We use an \textit{untied projection head} to predict the logits of relations in the mirrored relation predictor.
This means that the network weights between the mirrored and the original relation predictors are all tied except for the last prediction head.
Similar to other self-supervised learning works~\cite{chen2020simple,chen2020big}, we observe that having the untied projection head helps alleviate the burden of performing two tasks (\ie, the supervised relation prediction and the self-supervised relation alignment) for the original projection head, resulting in better learned representations. 

\subsection{Instantiation under SGTR}

To test the effectiveness of our proposed self-supervised relation alignment, we leverage it with a recently published Transformer-based SGTR model~\cite{li2022sgtr}. The choice of SGTR is motivated by two factors: (1) it is a state-of-the-art scene graph generation approach; and (2) its design mirrors DETR~\cite{carion2020end}, resulting in an one-stage end-to-end method.  

SGTR model~\cite{li2022sgtr} mainly contains four modules: (1) a {\em CNN feature extractor} with a multi-layer Transformer encoder to extract image features, (2) an {\em entity node generator} which predicts objects using a Transformer decoder with learned entity embedding queries conditioned on the image encoding, (3) a {\em predicate node generator} which predicts relations using a Transformer decoder with learned predicate embedding queries, and (4)
a {\em bipartite graph assembling module} for constructing the final bipartite matching between entities and 
predicates. This matching results in the final generated scene graph. 

The predicate node generator has two main components: (1) one {\em relation image feature encoder}, which refines the image feature extracted from the feature extraction part, specifically for relation prediction, and (2) a {\em structural predicate decoder}, which is a stack of self-attention and cross-attention Transformer blocks, 
to generate relation predictions. 
The structural predicate decoder receives two inputs, the refined image feature from the relation image feature encoder, and the entity feature 
from the entity node generator. 
It then makes three sets of predictions: one predicate/relation label distribution and the relation location, 
and two entity (the subject and the object related to the relation) label distributions and their locations. 

We instantiate our self-supervised relation alignment mechanism on the structural predicate decoder. 
Specifically, we first duplicate the structural predicate decoder and then apply our proposed masking on the input (image/entity feature) to the mirrored branch. 
The input to the structural predicate decoder is treated as {\tt key} and {\tt value} of its cross-attention Transformer blocks. 
Rather than applying the random masking on the input directly, inside a cross-attention Transformer block, the random mask is multiplied with the attention matrix in an element-wise fashion. 
In other words, each element in the attention matrix has an independent probability $p$ to be set to $-\mathrm{inf}$ before fed to the $\operatorname{softmax}$ operation. 
We generate a random mask per cross-attention Transformer block independently.

The alignment loss is enforced on the three label distributions predicted by the mirrored structural predicate decoder. 
Although the original structural predicate decoder makes predictions at each Transformer layer, we only apply the alignment loss at the last layer for simplicity. 
As described in Section~\ref{subsec:alignloss}, we untie the last projection layer of the mirrored predicate decoder with the original branch. 

\subsection{Instantiation under Neural Motifs}
\label{sec:model_motif}

To illustrate the generality of the self-supervised relational alignment, we also apply it to one of the most popular two-stage architectures for scene graph generation -- Neural Motifs~\cite{zellers2018neural}. 
Neural Motifs is built upon the Faster-RCNN~\cite{ren2015faster} object detector. It first utilizes a pre-trained object detector to extract input image features and generate object proposals. 
It then uses RoI Align operation to extract the object features and the relation features per pair of object proposals. 
Relying on these extracted features, it predicts a scene graph for a given input image.

Neural Motifs first inputs the object features and the object proposal location embeddings to a bi-directional LSTM (\emph{object context LSTM}) to predict object class labels. 
The predicted object labels, along with the object features, and the object context LSTM's hidden states, are then fed to another bi-directional LSTM (\emph{edge context LSTM}) to generate more refined object features. 
It then combines a pair of refined object features with the extracted relation feature and learned predicate prior (obtained from the empirical distribution of predicates in training data) to form a feature that is used to predict a relation label per pair of object proposals.

We apply our self-supervised relation alignment mechanism to the relation predictor. 
Specifically, we duplicate the edge context LSTM and its subsequent prediction layers. 
Random masks are applied on both input to the edge context LSTM and the relation feature. 
We replace the last two linear layers of the mirrored relation predictor with a three-layer MLP with ReLU activations. 
Similarly as before, this MLP is untied from the original relation predictor. 
In the mirrored predictor, the pairwise combined object features and the masked relation features are fed to the MLP which predicts the relation label distributions. 
The predicate prior is not used in the mirrored relation predictor.

The alignment loss is again calculated between the relation label distributions from the mirrored and the original relation predictors. 

\section{Experiments}
\label{sec:exper}
\noindent
\textbf{Dataset.} 
We conduct experiments on the widely used scene graph generation dataset, Visual Genome~\cite{krishna2017visual}. We use the same pre-processing procedure and train/val/test splits as previous works~\cite{xu2017scene,suhail2021energy,li2022sgtr,zellers2018neural}. 
This pre-processed Visual Genome dataset has $150$ object and $50$ relation categories.

\vspace{0.1in}
\noindent
\textbf{Evaluation Metrics.} 
We mainly use the mean-Recall@K (\textbf{mR@K}) metric to evaluate models, following previous works~\cite{chen2019knowledge,tang2019learning,tang2020unbiased}. Mean-Recall averages the Recall values computed for each relation class, which is a better evaluation metric than Recall, since it is less biased to dominant classes as suggested in~\cite{tang2020unbiased}. For a full comparison on the SGTR model, same as~\cite{li2022sgtr}, we also report the Recall@K (\textbf{R@K}) values and mR@100 for each long-tail category subset (\textbf{HEAD}, \textbf{BODY}, and \textbf{TAIL}), as proposed in~\cite{li2021bipartite}. The evaluation for the Neural Motifs model is conducted under three settings, (1) predicate classification (\textbf{PredCls}): predicting the relation labels given the image, ground-truth object bounding box locations and labels; (2) scene graph classification (\textbf{SGCls}): predicting both object labels and relation labels given the image and ground-truth bounding box locations; and (3) scene graph detection (\textbf{SGDet}): generating the whole scene graph based on the input image. 
Since there is no operation similar to RoI Align existing in the SGTR model, we only evaluate the SGTR model under the SGDet setting following~\cite{li2022sgtr}.

\subsection{SGTR with Relation Alignment}
\noindent
{\bf Implementation Details.} 
We use the codebase released by~\cite{li2022sgtr} to conduct the experiments. 
The pre-trained DETR~\cite{carion2020end}, learning rate scheduler, and all other hyper-parameter values are the same as those provided in the codebase. 
There are $\operatorname{dropout}$ operations throughout SGTR. 
For the SGTR model where our self-supervised relation alignment mechanism is applied, to eliminate the effect from the $\operatorname{dropout}$, 
we turn off all the $\operatorname{dropout}$ operations inside the mirrored structural predicate decoder, and we conduct another forward pass of the original decoder without the $\operatorname{dropout}$ operations to form the target of the alignment loss. 
We set the masking probability $p$ (in Equation~\ref{eq:mask_pattern}) to $0.1$ and the $\lambda$ (in Equation~\ref{eq:loss}) to $10$ for all three prediction alignments (the relation, the subject and the object related to the relation). 
Since our alignment mechanism introduces some randomness, we run both the original SGTR model and our self-supervised aligned SGTR variant $4$ times and report the average results along with the standard deviation. 
We select the best trained model for testing based on the mR@50 on the validation set. 

\begin{table*}[t]
  \centering
  \begin{tabular}{c|c|l|c|c|c|c|c|c|c}
    \toprule
    \textbf{B} & \textbf{D} & \textbf{Method} & \textbf{mR@50} & \textbf{mR@100} & \textbf{R@50} & \textbf{R@100} & \textbf{HEAD} & \textbf{BODY} & \textbf{TAIL} \\
    \midrule
    $\star$ & $\star$ & FCSGG~\cite{liu2021fully} & 3.6 & 4.2 & 21.3 & 25.1 & - & - & - \\
    \midrule
    \multirow{7}{*}{\rotatebox[origin=c]{90}{\bdsize X101-FPN}} & \multirow{8}{*}{\rotatebox[origin=c]{90}{\bdsize Faster RCNN}} & RelDN~\cite{zhang2019graphical} & 6.0 & 7.3 & 31.4 & 35.9 & - & - & - \\
    & & Motifs~\cite{tang2020unbiased} & 5.5 & 6.8 & \textbf{32.1} & \textbf{36.9} & - & - & - \\
    & & VCTree~\cite{tang2020unbiased} & 6.6 & 7.7 & 31.8 & 36.1 & - & - & - \\
    \cmidrule{3-10}
    & & VCTree-TDE~\cite{tang2020unbiased} & 9.3 & 11.1 & 19.4 & 23.2 & - & - & - \\
    & & VCTree-DLFE~\cite{chiou2021recovering} & 11.8 & 13.8 & 22.7 & 26.3 & - & - & - \\
    & & VCTree-EBM~\cite{suhail2021energy} & 9.7 & 11.6 & 20.5 & 24.7 & - & - & - \\
    & & VCTree-BPLSA~\cite{guo2021general} & 13.5 & 15.7 & 21.7 & 25.5 & - & - & - \\
    \cmidrule{1-1} \cmidrule{3-10}
    \multirow{6}{*}{\rotatebox[origin=c]{90}{\bdsize R101}} & & RelDN$^\ddag$~\cite{zhang2019graphical,li2022sgtr} & 4.4 & 5.4 & 30.3 & 34.8 & \textbf{31.3} & 2.3 & 0.0 \\
    \cmidrule{2-10}
    & \multirow{5}{*}{\rotatebox[origin=c]{90}{\bdsize DETR}} & AS-Net$^\ddag$~\cite{chen2021reformulating,li2022sgtr} & 6.1 & 7.2 & 18.7 & 21.1 & 19.6 & 7.7 & 2.7 \\
    & & HOTR$^\ddag$~\cite{kim2021hotr,li2022sgtr} & 9.4 & 12.0 & 23.5 & 27.7 & 26.1 & 16.2 & 3.4 \\
    & & SGTR~\cite{li2022sgtr} & 12.0 & 15.2 & 24.6 & 28.4 & 28.2 & 18.6 & 7.1 \\
    \cmidrule{3-10}
    & & SGTR* & 12.0 {\stdsize $\pm$ 0.4} & 16.0 {\stdsize $\pm$ 0.5} & 23.7 {\stdsize $\pm$ 0.4} & 26.9 {\stdsize $\pm$ 0.4} & 26.4 {\stdsize $\pm$ 0.3} & 20.0 {\stdsize $\pm$ 0.3} & 8.8 {\stdsize $\pm$ 0.8} \\
    & & Align-SGTR* & \textbf{12.7} {\stdsize $\pm$ 0.3} & \textbf{16.8} {\stdsize $\pm$ 0.2} & 24.6 {\stdsize $\pm$ 0.2} & 27.8 {\stdsize $\pm$ 0.2} & 27.3 {\stdsize $\pm$ 0.2} & \textbf{20.8} {\stdsize $\pm$ 0.1} & \textbf{9.8} {\stdsize $\pm$ 0.5} \\
    \bottomrule
  \end{tabular}
  \caption{\textbf{SGDet test results} for existing scene graph generation models on the Visual Genome dataset. Numbers are borrowed from~\cite{li2022sgtr} except for the last two rows. \textbf{B} means the backbone used for object detection, and \textbf{D} means the object detector type. $\star$ denotes the special backbone $\text{HRNetW48-5S-FPN}_{\times \text{2-f}}$ and object detector CenterNet~\cite{zhou2019objects} used in \cite{liu2021fully}. RelDN$^\ddag$, AS-Net$^\ddag$, and HOTR$^\ddag$ are trained by~\cite{li2022sgtr}. SGTR* is the original SGTR~\cite{li2022sgtr} model which we train and evaluate $4$ times using the authors' released code, and Align-SGTR* is the SGTR model equipped with our proposed self-supervised relation alignment mechanism during training, for which we also run $4$ times. Note that the structure and cost during inference for SGTR* and Align-SGTR* are identical; they only differ in training.}
  \label{tab:results_sgtr}
\end{table*}

\vspace{0.1in}
\noindent
{\bf Results.} 
Quantitative results are displayed in Table~\ref{tab:results_sgtr}. 
Our retrained SGTR model (denoted as SGTR*) and the SGTR variant with our proposed self-supervised relation alignment during training (denoted as Align-SGTR*) are listed at the last two rows of the table.  
To have a complete comparison with existing scene graph generation models, we also show the results of other competitors reported in~\cite{li2022sgtr}. 
There is some performance drop on Recalls comparing the results from the models trained using the DETR detector with those from the Faster RCNN detector. The main reason is that DETR is not as good as Faster RCNN in detecting small objects, while in the Visual Genome dataset more than half of the relations involve small objects; this was also discussed in \cite{li2022sgtr}. 

The results of our retrained SGTR model (SGTR*) match those reported in~\cite{li2022sgtr}. 
In particular, we obtain slightly higher BODY and TAIL results but slightly worse HEAD result compared to the reported, which results in slightly higher mean-Recalls, but slightly worse Recalls. 
This is reasonable, because it is now well-known that for a specific model, increase in mean-Recalls often leads to decrease in Recalls~\cite{lu2021context,tang2020unbiased,zareian2020bridging}. 

Comparing the results between SGTR* and our Align-SGTR*, it is clear that the self-supervised alignment provides consistent and significant (at least two standard deviation higher) improvements on \textit{all} evaluation metrics. 
This demonstrates the effectiveness of our proposed mechanism. 
Our proposed self-supervised relation alignment is simple; it only introduces the extra network weights for the additional untied projection head 
and a KL loss during training, but very effective. 
By comparing the results between Align-SGTR* and all other scene graph generation models in the table, 
Align-SGTR* achieves the best mean-Recall, BODY, and TAIL results. 

\subsection{Neural Motifs with Relation Alignment}
\noindent
{\bf Implementation Details.} 
Neural Motifs, as other two-stage scene graph generation models, requires pre-training an object detector. 
We first train a Faster RCNN object detector with VGG-16 backbone using the codebase provided by~\cite{tang2020unbiased}, and then train the Neural Motifs model based on that. 
While training the Neural Motifs model, we use a batch size of $24$, and SGD optimizer with momentum $0.9$ and weight decay $0.0001$. 
We set the initial learning rate to $0.01$ and use the cosine annealing schedule~\cite{loshchilov2016sgdr} to decay the learning rate during training, with minimum learning rate set to $0$. 
Learning rate is updated every $1$K iterations and we train the model for a total of $100$K iterations. 
We select the best trained model for testing based on the validation mR@50. 
For our self-supervised relation alignment, we set the masking probability $p$ (in Equation~\ref{eq:mask_pattern}) to $0.1$ and the $\lambda$ (in Equation~\ref{eq:loss}) to $10$. The alignment loss is only added for the relation prediction. 
We run the training of both the Neural Motifs model and our self-supervised relation aligned variant $4$ times. 

\begin{table*}[t]
  \centering
  \small
  \begin{tabular}{l|c|c|c|c|c|c|c|c|c}
    \toprule
    \textbf{Method} &  \multicolumn{3}{c|}{\multirow{2}{*}{\textbf{PredCls}}} & \multicolumn{3}{c}{\multirow{2}{*}{\textbf{SGCls}}} & \multicolumn{3}{|c}{\multirow{2}{*}{\textbf{SGDet}}} \\
    \cmidrule{1-1}
    \textbf{B}: VGG-16 &  \multicolumn{3}{c|}{} & \multicolumn{3}{c}{} & \multicolumn{3}{|c}{} \\
    \textbf{D}: Faster RCNN & \textbf{mR@20} & \textbf{mR@50} & \textbf{mR@100} & \textbf{mR@20} & \textbf{mR@50} & \textbf{mR@100} & \textbf{mR@20} & \textbf{mR@50} & \textbf{mR@100} \\
    \midrule
    IMP~\cite{xu2017scene} & - & 9.8 & 10.5 & - & 5.8 & 6.0 & - & 3.8 & 4.8 \\
    Motifs~\cite{zellers2018neural} & 10.8 & 14.0 & 15.3 & 6.3 & 7.7 & 8.2 & 4.2 & 5.7 & 6.6 \\
    VCTree~\cite{tang2019learning} & 14.0 & 17.9 & 19.4 & \textbf{8.2} & 10.1 & 10.8 & \textbf{5.2} & \textbf{6.9} & \textbf{8.0} \\
    \midrule
    Motifs$^\ddag$~\cite{zellers2018neural,khandelwal2021segmentation} & 13.7 & 17.5 & 18.9 & 7.5 & 9.2 & 9.8 & \textbf{5.2} & 6.8 & 7.9 \\
    \midrule
    Motifs* & 15.0 {\stdsize $\pm$ 0.8} & 19.8 {\stdsize $\pm$ 0.9} & 21.9 {\stdsize $\pm$ 0.9} & 7.2 {\stdsize $\pm$ 0.3} & 8.9 {\stdsize $\pm$ 0.4} & 9.6 {\stdsize $\pm$ 0.4} & 4.2 {\stdsize $\pm$ 0.4} & 5.9 {\stdsize $\pm$ 0.5} & 7.2 {\stdsize $\pm$ 0.3} \\
    Align-Motifs* & \textbf{16.5} {\stdsize $\pm$ 0.4} & \textbf{21.7} {\stdsize $\pm$ 0.5} & \textbf{24.2} {\stdsize $\pm$ 0.5} & \textbf{8.2} {\stdsize $\pm$ 0.2} & \textbf{10.5} {\stdsize $\pm$ 0.2} & \textbf{11.5} {\stdsize $\pm$ 0.3} & 4.4 {\stdsize $\pm$ 0.2} & 6.4 {\stdsize $\pm$ 0.3} & 7.8 {\stdsize $\pm$ 0.3} \\
    \bottomrule
  \end{tabular}
  \caption{\textbf{PredCls, SGCls, SGDet test results} on the Visual Genome dataset for models with VGG-16 backbone (\textbf{B}) and Faster-RCNN detector (\textbf{D}). Numbers are borrowed from~\cite{khandelwal2021segmentation} except for the last two rows. Motifs$^\ddag$ is trained by~\cite{khandelwal2021segmentation}. Motifs* is our trained Neural Motifs~\cite{zellers2018neural} model, and Align-Motifs* is the Neural Motifs model containing our proposed self-supervised relation alignment mechanism during training, and for both we train the models $4$ times.}
  \label{tab:results_motif}
\end{table*}

\vspace{0.1in}
\noindent
{\bf Results.} 
Quantitative results are shown in Table~\ref{tab:results_motif}. 
The results of our retrained Neural Motifs model (denoted as Motifs*) and the variant with our self-supervised relation alignment (denoted as Align-Motifs*) are listed at the last two rows of the table.  
We again report the mean and the standard deviation of the results collected from $4$ runs.
We also copy the results of other scene graph generation models from~\cite{khandelwal2021segmentation}.  
All these competitors use Faster~RCNN (with VGG-16 backbone) as the object detector.
The results of our re-trained Motifs* roughly match those of Motifs$^\ddag$ provided by~\cite{khandelwal2021segmentation}. 
The main discrepancy of the results is caused by the fact that we use a different codebase to train the Faster RCNN object detector than theirs.

From Table~\ref{tab:results_motif}, we can see that our Align-Motifs* achieves significantly better (larger than at least one standard deviation) results than those of Motifs* on \textit{all} metrics. 
Moreover, under the PredCls and SGCls settings, our Align-Motifs* achieves the best results among all the (Faster RCNN with VGG-16 backbone) models. 
The improvements of Align-Motifs* over Motifs* across the three evaluation settings are PredCls $>$ SGCls $>$ SGDet. 
This is reasonable, because our proposed self-supervised alignment mechanism is only applied to relation prediction. 
Applying a similar self-supervised alignment mechanism to the object prediction part could potentially further improve the performance under the SGCls and SGDet settings. 
One possible reason for the marginal improvement under the SGDet setting is that the pre-trained Faster RCNN 
is fixed during Neural Motif's training. 
If the object proposals are not good (\eg, the detection recall is low), then the performance of the scene graph generation model will be significantly reduced. 
This is an inherent limitation of a two-stage scene graph generation model.

\begin{figure*}[t]
  \centering
  \includegraphics[width=\linewidth]{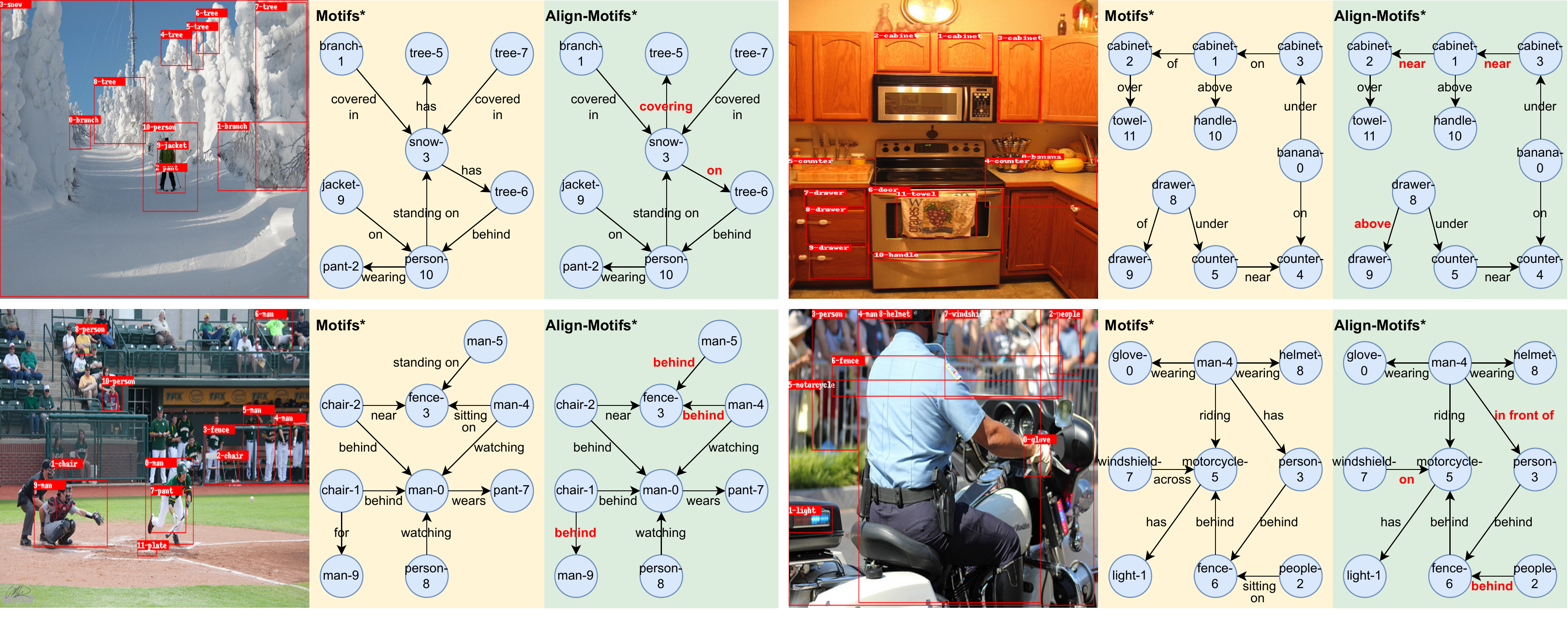}
  \caption{{\bf Qualitative Results.} Relation prediction results from Motifs* and Align-Motifs* trained under the PredCls setting.
  }
  \label{fig:quali}
  \vspace{-0.09in}
\end{figure*}

We visualize some qualitative results 
for relation prediction in Figure~\ref{fig:quali}. These results are obtained from Motifs* and Align-Motifs* models trained under the PredCls setting. The qualitative results confirm the performance gain shown in the quantitative results. The Align-Motifs* model generally generates more accurate relation predictions than the Motifs* model.

\subsection{Ablation Study}
We conduct an ablation study based on the SGTR model to verify the effectiveness of our design choices: (1) our self-supervised alignment vs. a supervised alignment, and (2) the untied projection head vs. a shared projection head in the mirrored relation predictor. 
Experimental results are listed in Table~\ref{tab:results_sgtr_ab}. 
In the table, the performance of the baseline SGTR is our reproduced result. 
These are validation results obtained from a single run on the same machine with the same random seed.

\begin{table}[t]
  \centering
  \resizebox{\linewidth}{!}{
  \begin{tabular}{l|c|c|c|c}
    \toprule
    \textbf{Method} & \textbf{mR@50} & \textbf{mR@100} & \textbf{R@50} & \textbf{R@100} \\
    \midrule
    SGTR & 14.9 & 18.4 & 24.0 & 27.4 \\
    SGTR + SA  + UPH & 15.0 & 18.6 & 24.5 & 27.8 \\
    SGTR + SSA + PH & 15.9 & 19.2 & 24.6 & \textbf{27.9} \\
    SGTR + SSA + UPH & \textbf{16.0} & \textbf{20.5} & \textbf{24.8} & \textbf{27.9} \\
    \bottomrule
  \end{tabular}
  }
  \caption{\textbf{Ablation study with validation results.} SGTR is retrained by us. Self-supervised alignment (SSA) has one supervised loss in the original predictor and a KL loss between the original and the mirrored predictors. Supervised alignment (SA) has two supervised losses on the original and the mirrored predictors respectively. PH/UPH denotes the tied/untied projection head in the mirrored relation predictor. 
  }
  \label{tab:results_sgtr_ab}
\end{table}

\vspace{0.1in}
\noindent
{\bf Self-Supervised vs. Supervised Alignment.} 
We first compare our self-supervised alignment loss to a supervised one. 
Particularly, we construct a supervised relation alignment module by following the exact same setting as the self-supervised module except that we remove the KL loss and plug the same supervised loss in the mirrored relation predictor. 
This supervised loss will push the mirrored predictor towards predicting the correct relation labels (\ie, aligning with the ground truth labels) whereas the self-supervised loss will push it to align with the original predictor. 
By switching our self-supervised loss to a supervised one, the mean-Recall results are reduced by almost $1$ point. 
Due to the random masking applied on the input, placing a supervised loss on the mirrored branch only slightly improves the robustness of the learned representations.
However, our self-supervised loss acts differently.
It is like the distillation or a student-teacher model, \ie, the mirrored predictor learns to mimic the original one. 
This provides a more informative label distribution supervision, \eg, with a higher entropy compared to the one-hot ground truth labels (Dirac delta distribution). 
It can also provide some supervision on unlabeled relation examples, which often appears as the ground truth relation labels are incomplete and very long-tail distributed. 

\vspace{0.1in}
\noindent
{\bf Untied vs. Tied Projection Head.} 
We then validate the effect of the untied projection head vs. a tied projection head in the mirrored relation predictor, where the weights of the relation logits predictor are also shared between the mirrored and the original predictors. 
As can be seen, though the self-supervised alignment loss alone already brings a significant benefit, the untied projection head further improves the performance. 
Adding this extra projection head helps alleviate the burden of performing two tasks (\ie, the supervised relation prediction and the self-supervised relation alignment) simultaneously, resulting in better learned representations. 
Similar observation has been made in other self-supervised learning works~\cite{chen2020simple,chen2020big}. 

\section{Conclusion}
\label{sec:conclu}
We propose a simple-yet-effective self-supervised relational alignment module that can be plugged into any existing scene graph generation model as an additional loss term.
Particularly, we mirror the relation prediction branch and feed randomly masked input to the mirrored branch. 
The alignment between the predictions from the mirrored and the original branches encourages the model to learn better representations. 
Experiments with two 
scene graph models (one one-stage, and one two-stage) on the Visual Genome dataset show that our method significantly improves performance.
In the future, we plan to design more sophisticated random masking patterns and explore this method in the setting where a larger scale of unsupervised data is available.

\vspace{0.06in}
\noindent
\textbf{Discussion on potential negative societal impact.} The scene graph generation task is 
for better representation of a given input image. The task itself and the models built for the task are neutral; potential negative societal impact may be introduced in how human beings using this representation. 
The method proposed in our work aims to improve the performance of existing models on this specific task. We do not introduce any further potential negative societal impact. 

\vspace{0.06in}
\noindent 
\textbf{Acknowledgments.} 
This work was funded, in part, by the Vector Institute for AI, Canada CIFAR AI Chairs, NSERC CRC, and NSERC DGs. Resources used in preparing this research were provided, in part, by the Province of Ontario, the Government of Canada through CIFAR, the Digital Research Alliance of Canada \url{alliance.can.ca}, \href{https://vectorinstitute.ai/\#partners}{companies} sponsoring the Vector Institute, and Advanced Research Computing at the University of British Columbia. Additional hardware support was provided by John R. Evans Leaders Fund CFI grant and Compute Canada under the Resource Allocation Competition award.

{\small
\bibliographystyle{ieee_fullname}
\bibliography{egbib}
}

\clearpage
\maketitlesupplementary

\section{Hyperparameters}
\subsection{Model Performance w.r.t. the Masking Ratio}
We conduct ablation experiments with respect to the masking ratio $p$ (in Equation~\ref{eq:mask_pattern} of the main paper) on the Neural Motifs~\cite{zellers2018neural} model under the PredCls setting. 
Table~\ref{tab:motif_p} below shows the validation mean-Recall results with different $p$ values, which are obtained from a single run on the same machine with the same random seed. The loss weight $\lambda$ (in Equation~\ref{eq:loss} of the main paper) is fixed to $10$ for all the experiments.
As can be seen, $p$ being $0.1$ gives us the best validation results among the values we tried. Based on this ablation, we set $p$ to be $0.1$ for other Neural Motif’s training settings (SGCls and SGDet), and for SGTR. Notably, even with a sub-optimal $p$ value, over a wide range of $p$ values, we obtain significant improvements over the Motifs baseline. 

\begin{table}[h]
  \centering
  \begin{tabular}{l|c|c}
    \toprule
    \textbf{Method} & \textbf{mR@50} & \textbf{mR@100} \\
    \midrule
    Motifs & 22.1 & 24.2 \\
    Align-Motifs ($p=0.05$) & 23.7 & 25.9 \\
    Align-Motifs ($p=0.1$) & \textbf{23.9} & \textbf{26.2} \\
    Align-Motifs ($p=0.2$) & 23.5 & 25.8 \\
    Align-Motifs ($p=0.4$) & 23.2 & 25.4 \\
    Align-Motifs ($p=0.6$) & 22.8 & 24.8 \\
    \bottomrule
  \end{tabular}
  \caption{\textbf{Validation results of different masking ratio $p$ values for the Neural Motifs model under the PredCls setting.} Motifs is our trained Neural Motifs~\cite{zellers2018neural} model. Align-Motifs is the Neural Motifs model containing our proposed self-supervised relation alignment mechanism during training, where the $p$ value in the bracket is the masking ratio used for the experiment.
  }
  \label{tab:motif_p}
\end{table}

\subsection{Model Performance w.r.t. the Loss Weight}
Again, on the Neural Motifs~\cite{zellers2018neural} model under the PredCls setting, we conduct ablation experiments with respect to the loss weight $\lambda$ (in Equation~\ref{eq:loss} of the main paper). 
Table~\ref{tab:motif_lambda} below shows the validation mean-Recall results with different $\lambda$ values, which are obtained from a single run on the same machine with the same random seed. The masking ratio $p$ (in Equation~\ref{eq:mask_pattern} of the main paper) is fixed to $0.1$ for all the experiments.
As the results suggest, $\lambda$ being $10$ gives us the most effective validation results among the values we experimented. Based on this ablation, we set $\lambda$ to be $10$ for all other experiment settings.

\begin{table}[ht]
  \centering
  \begin{tabular}{l|c|c}
    \toprule
    \textbf{Method} & \textbf{mR@50} & \textbf{mR@100} \\
    \midrule
    Motifs & 22.1 & 24.2 \\
    Align-Motifs ($\lambda=0.1$) & 22.0 & 23.5 \\
    Align-Motifs ($\lambda=1$) & 23.8 & 25.9 \\
    Align-Motifs ($\lambda=10$) & \textbf{23.9} & \textbf{26.2} \\
    Align-Motifs ($\lambda=50$) & 22.5 & 24.2 \\
    Align-Motifs ($\lambda=100$) & 21.7 & 23.4 \\
    \bottomrule
  \end{tabular}
  \caption{\textbf{Validation results of different loss weight $\lambda$ values for the Neural Motifs model under the PredCls setting.} Motifs is our trained Neural Motifs~\cite{zellers2018neural} model. Align-Motifs is the Neural Motifs model containing our proposed self-supervised relation alignment mechanism during training, where the $\lambda$ value in the bracket is the loss weight used for the experiment.
  }
  \label{tab:motif_lambda}
\end{table}

\section{Masking Illustration - Align-SGTR*}
\begin{figure*}[t]
  \centering
  \includegraphics[width=\linewidth]{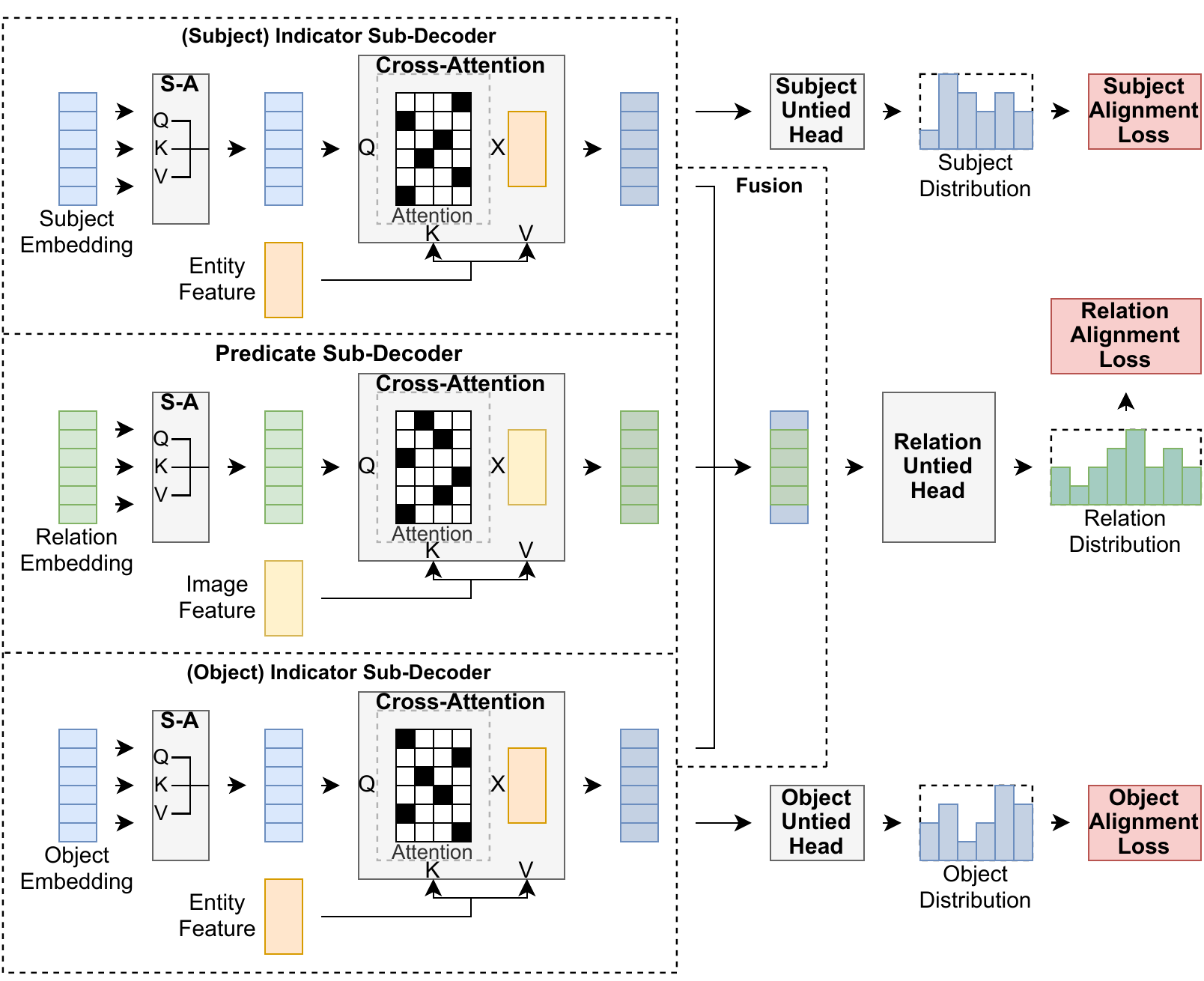}
  \caption{{\bf Instantiation of our proposed self-supervised relation alignment mechanism under SGTR.} This figure illustrates the last Transformer layer of the mirrored structural predicate decoder, which shares weights with the original one, except for the untied projection heads ({\tt Relation/Subject/Object Untied Head}). Random masking is applied on the attention matrix of the cross-attention Transformer blocks. Mask is generated independently for every cross-attention Transformer block at each Transformer layer, while the alignment losses are only enforced at the last layer. In the figure, {\tt S-A} stands for a self-attention Transformer block. {\tt Q}, {\tt K}, and {\tt V} are the {\tt query}, {\tt key}, and {\tt value} of a Transformer block respectively.}
  \label{fig:mask-sgtr}
\end{figure*}

Figure~\ref{fig:mask-sgtr} illustrates the last layer of the mirrored relation predictor (mirrored structural predicate decoder) of Align-SGTR* -- the SGTR~\cite{li2022sgtr} model equipped with our proposed self-supervised relation alignment mechanism during training. The mirrored structural predicate decoder shares weights with the original one, except for the untied projection heads. Random masking is applied on the attention matrix of the cross-attention Transformer blocks. Mask is generated independently for every cross-attention Transformer block at each Transformer layer, while the alignment losses are only enforced at the last layer.

\section{Per-Predicate R@100 Difference - SGTR}
\begin{figure*}[t]
  \centering
  \includegraphics[width=\linewidth]{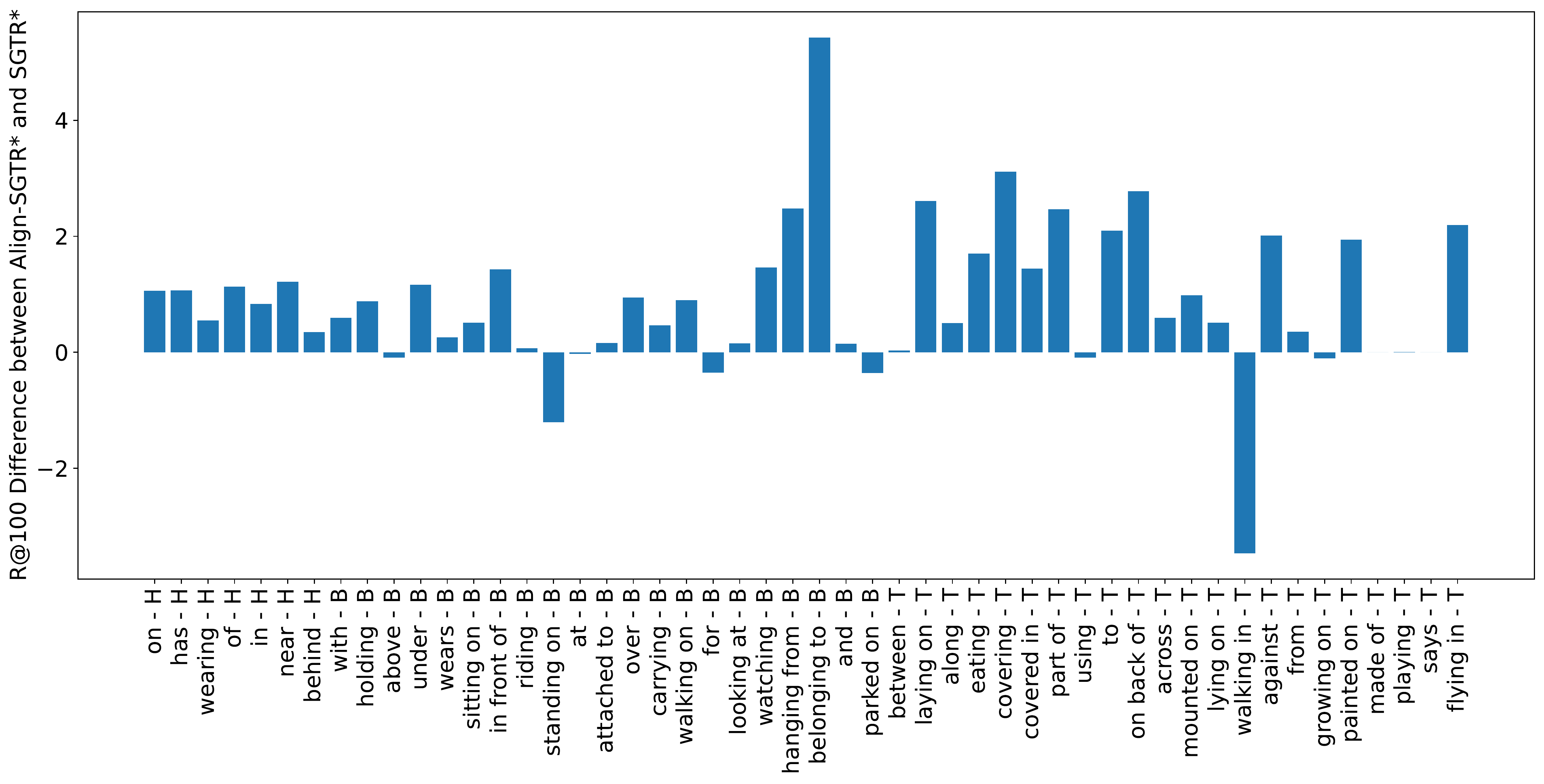}
  \caption{{\bf Per-predicate R@100 difference between Align-SGTR* and SGTR*.} The predicates are sorted by their frequencies in descending order from left to right. {\tt H}, {\tt B}, and {\tt T} indicate the head, body, and tail partitions respectively. Results are averaged across $4$ runs.}
  \label{fig:diff-sgtr}
\end{figure*}

Figure~\ref{fig:diff-sgtr} shows the per-predicate R@100 difference between Align-SGTR* and SGTR* (the SGTR~\cite{li2022sgtr} model trained by us). The predicates are sorted by their frequencies in descending order from left to right. The predicate order, and the head-body-tail partitions are from~\cite{li2021bipartite}. Results are averaged across $4$ runs. As can be seen, our Align-SGTR* is better than SGTR* on $40$ predicate labels (out of a total of $50$), in many cases by a sizable margin. 

\end{document}